\documentclass[sigconf, nonacm]{acmart} 
\renewcommand\footnotetextcopyrightpermission[1]{} 

\usepackage{graphicx}
\usepackage{libertine}
\usepackage{array}
\usepackage{float}
\usepackage{tikz}
\usepackage{multirow}
\usepackage{multicol}
\usepackage{makecell}
\usepackage{hyperref}
\usepackage{booktabs}

\AtBeginDocument{%
  \providecommand\BibTeX{{%
    \normalfont B\kern-0.5em{\scshape i\kern-0.25em b}\kern-0.8em\TeX}}}

\begin{document}

\title{\textsf{LimTopic: LLM-based Topic Modeling and Text Summarization for Analyzing Scientific Articles limitations}}


\author{\textsf{Ibrahim Al Azher}}
\affiliation{%
  \institution{Northern Illinois University}
  \city{Dekalb}
   \city{Dekalb, IL}
   \country{USA}
  }
\email{iazher1@niu.edu}

\author{\textsf{Venkata Devesh Reddy}}
\affiliation{%
  \institution{Northern Illinois University}
  \city{Dekalb, IL}
  \country{ USA}
}
\email{dseethi@niu.edu}

\author{\textsf{Hamed Alhoori}}
\affiliation{%
 \institution{Northern Illinois University}
  \city{Dekalb, IL}
  \country{ USA}
  }
\email{alhoori@niu.edu}

\author{\textsf{Akhil Pandey Akella}}
\affiliation{%
 \institution{Northwestern University}
   \city{Evanston, IL}
   \country{USA}
   }
\email{akhilpandey.akella@kellogg.northwestern.edu}

\renewcommand{\shortauthors}{I.A Azher et al.}

\begin{abstract}
  The ``limitations'' sections of scientific articles play a crucial role in highlighting the boundaries and shortcomings of research, thereby guiding future studies and improving research methods. Analyzing these limitations benefits researchers, reviewers, funding agencies, and the broader academic community. We introduce \textbf{LimTopic}, a strategy where \textbf{Topic} generation in \textbf{Lim}itation sections in scientific articles with Large Language Models (LLMs). Here, each topic contains the title and `Topic Summary.' This study focuses on effectively extracting and understanding these limitations through topic modeling and text summarization, utilizing the capabilities of LLMs. We extracted limitations from research articles and applied an LLM-based topic modeling integrated with the BERtopic approach to generate a title for each topic and `Topic Sentences.' To enhance comprehension and accessibility, we employed LLM-based text summarization to create concise and generalizable summaries for each topic's Topic Sentences and produce a `Topic Summary.' Our experimentation involved prompt engineering, fine-tuning LLM and BERTopic, and integrating BERTopic with LLM to generate topics, titles, and a topic summary. We also experimented with various LLMs with BERTopic for topic modeling and various LLMs for text summarization tasks. Our results showed that the combination of BERTopic and GPT 4 performed the best in terms of silhouette and coherence scores in topic modeling, and the GPT4 summary outperformed other LLM tasks as a text summarizer. Our code and dataset are available at 
  \url{https://github.com/IbrahimAlAzhar/LimTopic/tree/master}. 
\end{abstract}



\maketitle

\section{\textsf{Introduction}}
Examining the limitations sections of research articles offers numerous benefits to researchers, institutions, and policymakers \cite{hyder2011national}. This analysis helps researchers learn from past shortcomings, identify new research avenues, and set realistic project expectations, ultimately fostering a more productive research environment across diverse fields \cite{hennink2005using}. Early-career researchers, in particular, can accelerate their learning by studying these sections \cite{nicholas2015younger}. They can develop novel approaches that advance their fields by uncovering potential pitfalls and promising research gaps. Experienced researchers can also gain deeper insights into their own research limitations, which may lead to methodological innovations and the development of new approaches that overcome existing shortcomings \cite{aguinis2018you}. Recognizing these limitations can also catalyze cross-institutional and interdisciplinary collaborations to tackle complex issues \cite{hara2003emerging,freeman2020measuring,9689840}.

Furthermore, Editors and reviewers can expedite the review process by understanding the drawbacks of a particular field. For institutions and funding agencies \cite{thelwall2023research} such as NSF, NIH, and DoE, understanding research limitations could support the strategic allocation of resources toward areas needing further exploration \cite{yu2021demand,ortagus2020performance,conaway2015research}. Additionally, policymakers can play a crucial role in addressing identified limitations by implementing supportive policies and providing targeted funding \cite{pielke2007honest,oliver2019understanding,altmann2020policy,o1986policy}. Overall, a thorough analysis of limitations sections promotes transparency \cite{ioannidis2005most,haibe2020transparency}, facilitates collaboration, and ultimately enhances the quality and impact of future research initiatives \cite{glass1976primary}. Researchers can utilize this structured categorization to better understand current limitations in their field and direct future studies accordingly. This enhances the accessibility and usefulness of the information for researchers and practitioners \cite{booth2021systematic}.

In this research, we propose a pipeline to process limitations sections from research papers and produce a list of topics with clear, accessible descriptions. Our approach uses an LLM-based topic modeling technique to generate insightful topics and concise summaries that capture the essence of limitations discussed in research articles. We experimented with models like Latent Dirichlet Allocation (LDA) and BERTopic \cite{grootendorst2022bertopic}, as well as various LLMs, including GPT-4, GPT-3.5 \cite{brown2020language}, Mistral, and Llama 2, utilizing different prompting methods. Each generated topic is titled with a meaningful label (Table \ref{tab:title-topic}). 
We applied BERTopic with LLMs to generate topics with titles, and BERTopic generated Topic Sentence for each topic by analyzing the topics (clusters) and selecting the most central documents for each topic. The Topic Sentences for each topic comprised multiple lengthy, complex sentences that were difficult to interpret because it collected various sentences from the dataset where some sentences were too specific for particular limitations. To address this issue, we implemented a text summarization approach using LLMs to create concise summaries for each topic, enhancing their comprehensibility. We experimented with various LLMs for text summarization, including GPT-4 \cite{brown2020language}, Llama 3, Claude 3.5 Sonnet, and GPT-3.5. 





The key contributions of this work are as follows:

\begin{enumerate}
    \item We developed a structured approach to categorize research limitations through topic modeling and to identify key themes within the limitations.
    \item We integrated LLMs with BERTopic for topic modeling and generating titles for each topic. We also applied LLMs to create easy-to-understand, concise, and generalizable summaries for each topic by distilling the key details of each topic.
    \item We conducted experiments and comparative analysis using various LLMs to determine the optimal performance for topic modeling and text summarization and evaluated the results.
\end{enumerate}


%

\section{\textsf{Related Work}}

Scholarly articles have long been recognized as a rich source for diverse types of information \cite{williams2014scholarly}. Over the years, numerous studies have developed sophisticated tools and methods for extracting elements such as abstracts  \cite{wu2015pdfmef}, headers, keyphrases, methodologies, figures, tables, acknowledgments \cite{khabsa2012ackseer}, future work, and references from PDF documents. Initial approaches often combine rule-based methods with machine learning techniques to enhance the accuracy and scope of extraction. For instance, \citet{houngbo2012method} focused on extracting methodological sections using a rule-based approach supplemented by machine learning. Further developments saw the integration of specific tools to facilitate the extraction process. Recent advancements have utilized more sophisticated algorithms and deep learning such as used an unsupervised graph-based algorithm for keyphrase extraction \cite{patel2021exploiting}. The use of Convolutional Neural Networks (CNNs) marked a significant evolution, as demonstrated by \citet{ling-chen-2020-deeppapercomposer}, who applied CNNs to extract both textual and non-textual content, showcasing the potential of deep learning in managing complex document structures. Complementarily, \citet{oelen2020generate} focused on comparing and aligning similar research contributions, assisting researchers in navigating prevalent research themes. 

Topic modeling techniques, such as Latent Dirichlet Allocation (LDA), have been widely used across various fields \cite{song2009topic}. While LDA treats documents as a bag of words, it has notable limitations, including the need to specify a fixed number of topics in advance and its sensitivity to hyperparameters. To enhance the coherence of topic representations, a bidirectional transformer model, BERT \cite{devlin2018bert}, can be used, which captures the context of a word by considering surrounding words. However, even with these improvements, interpreting topics can be challenging when relying solely on keywords or topic words. To address this, we incorporate large language models (LLMs) such as GPT with BERTopic \cite{grootendorst2022bertopic} to generate meaningful titles for each topic, making them more comprehensible than word representations alone.

The application of Large Language Models (LLMs) to text summarization has significantly transformed the process, often yielding results that surpass human-generated summaries in terms of efficiency and cost-effectiveness \cite{pu2023summarization, chang2023booookscore}. For example, the Long T5 model has proven effective for summarizing multiple documents in literature reviews \cite{yu-2022-evaluating}. Similarly, BERT has been tailored to summarize scientific texts \cite{gupta-etal-2021-effect}, and newer models \cite{varab2023abstractive} are capable of producing both abstractive and extractive summaries within a unified framework. LLMs have also been used to facilitate topic generation by extracting and aggregating topics from sentences within documents. This process involves condensing the extracted topics to provide a concise overview \cite{10386113}. Further extending this application, BERTopic has been integrated with LLMs for specific uses: \citet{koloski2024aham} applied BERTopic with Llama 2 for literature mining, and \citet{kato2024u} utilized it alongside GPT for quantitative political analysis. These efforts have been enhanced by fine-tuning LLMs to optimize topic modeling capabilities \cite{kajoluoto2024internet}. 
Despite these advancements, the field of extracting and generating limitations from scientific articles remains relatively underexplored. While there have been initiatives like those by and \citet{faizullah2024limgen} to generate limitations from scientific texts, and efforts to apply multimodal LLMs for this purpose \cite{10826112}, the specific application of BERTopic combined with LLMs to model topics and summarize limitation sections in scientific papers has not been widely pursued. This gap highlights a critical area for future research, presenting opportunities to enhance the depth and accuracy of scientific literature analysis.

\begin{figure}[ht]
    \includegraphics[width=\columnwidth,keepaspectratio]{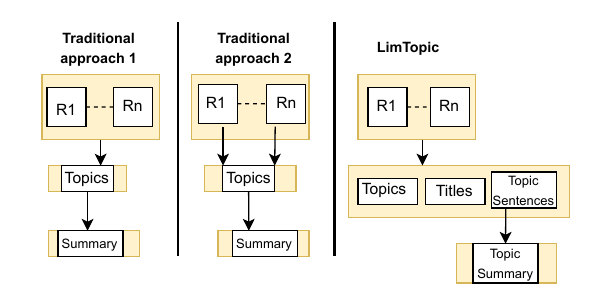}

\begin{tabular}{|l|p{1.5cm}|p{1.5cm}|p{1.5cm}|}   
\hline
\textbf{Feature} & 
\textbf{\makecell{Traditional \\ Approach 1}}
 & \textbf{\makecell{Traditional \\ Approach 2}} & \textbf{LimTopic} \\
\hline
Global Context & $\bullet$ &  & $\bullet$ \\
Local Context &  & $\bullet$ & $\bullet$ \\
Contextual Details &  &  & $\bullet$ \\
Readable Title &  &  & $\bullet$  \\
\hline
\end{tabular}
\caption{Traditional approaches vs. LimTopic.}
\label{tab:other_vs_ours_table}
\Description{Comparing features of LimTopic with traditional approaches for text summarization.}
\vspace{-13pt}
\end{figure}



\begin{figure}
\includegraphics[width=\columnwidth, keepaspectratio]{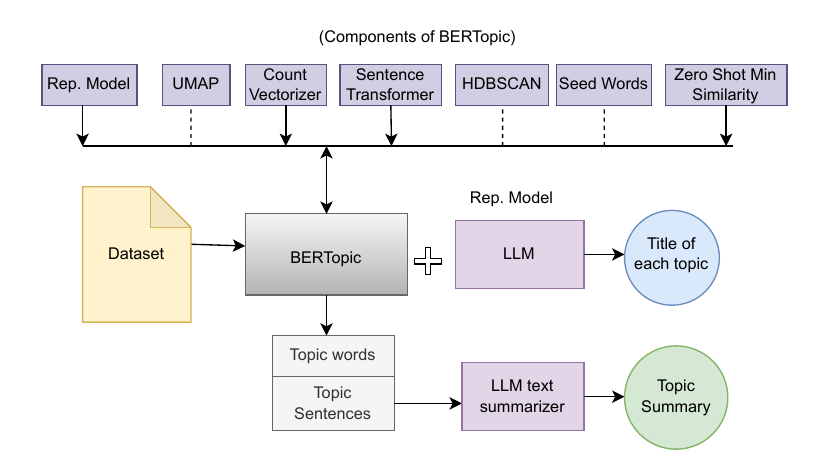}
  \caption{Details about BERTopic + LLM for generating title, topic, and Topic Summary.}
  \label{fig:BERTopic}
  \Description{Details about BERTopic + LLM for generating title, topic, and Topic Summary.}
  \vspace{-16pt}
\end{figure}

Figure \ref{tab:other_vs_ours_table}  illustrates the differences between our approach, LimTopic, and traditional methods. Here, R denotes a research paper, and 
n is the dataset's total number of research papers. Traditional summarization methods typically follow one of two approaches: \\
\textbf{Traditional Approach 1:} This approach combines all research papers into a single dataset and applies a summarization algorithm to create topics and summaries. While it preserves global context, it often loses local context and relevant information specific to individual papers. \\
\textbf{Traditional Approach 2:} This approach generates individual summaries for each paper and then aggregates them. Although it maintains local context, it struggles to capture the global context, often resulting in redundant information and lacking cohesive insights across documents.
To address these limitations, our method, LimTopic, synthesizes both global and local contexts. We use BERTopic with LLMs to create cohesive topics and generate Topic Sentences from the entire dataset, ensuring global context and relational coherence across papers. Next, we apply LLM-based text summarization to refine each topic’s Topic Sentences, producing concise summaries that capture local context and maintain relevant details. This process not only preserves contextual depth but also generates meaningful, clear topic titles that offer cohesive insights across all research papers.

\begin{figure*}
  \includegraphics[scale=0.80]{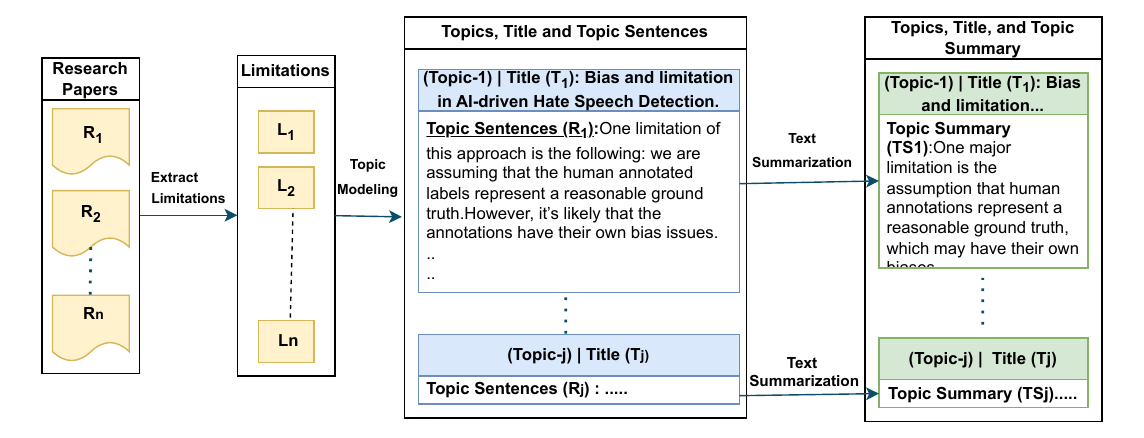}
  \caption{LimTopic for generating meaningful topic titles with summaries from a large collection of text-form limitations
sections.} 
  \Description{LimTopic for generating meaningful topic titles with summaries from a large collection of text-form limitations
sections.}
  \label{fig:archi-diagram}
  \vspace{-11pt}
\end{figure*}

\section{\textsf{Dataset Collection and Pre-processing}}


\textbf{Data collection:} We took ACL research papers from the year 2023, which comprise 2,896 papers consisting of 231 short papers, 1261 long papers, 976 findings papers, and 428 papers from the remaining 24 categories. We extracted all of the `limitation' sections using a hybrid approach. We used the Science Parse\footnote{https://github.com/allenai/science-parse} tool to extract text in a JSON format, where section titles (e.g., abstract, introduction, limitations) serve as keys and text of each section as the value. At first, we applied this tool to automatically detect limitations sections from scientific papers. We collected 1,406 papers with limitations sections. However, Science Parse did not detect many limitations sections because they were sub-sections within other sections. 


The Science Parse tool can detect a section as a separate section if any number (e.g., 1, 2, 3) appears as a prefix in a section heading; otherwise, it will treat it as a sub-section within the other section. So, we applied a rule-based approach to detect the rest of the 1,490 limitations sections. We searched for the word `limitation' or `limitations' in the ethics, conclusion, future work, broader impact, discussion, and other sections, except the abstract, introduction, related work, methodology, and acknowledgments sections since we found that these sections mostly didn't discuss their limitations, and sometimes, they discussed other papers' limitations. We initiated the extraction process when we encountered the word `limitation' or `limitations' and continued until we reached the beginning of the subsequent section or sub-section.

\begin{table}[h!t]
  \centering
  \begin{tabular}{|>{\raggedright\arraybackslash}m{1.9cm}|m{2.75cm}|m{2.75cm}|}
    \hline
    \textbf{Categories} & \textbf{Explicit limitations} & \textbf{Implicit limitations} \\ \hline
   Short Papers & 81 & 84 \\ \hline
      Long Papers & 368 & 544 \\ \hline
      Findings ACL & 573 & 510 \\ \hline
      Semantic Eval. & 281 & 38 \\ \hline
      Industry Track & 46 & 31 \\ \hline
  \end{tabular}
  \caption{Number of papers with Explicit and Implicit limitations sections.}
  \Description{Number of papers with Explicit and Implicit limitations sections.}
  \label{tab:paper-statistics}
  \vspace{-12pt}
\end{table}

\begin{table}[!ht]
  \centering
  \begin{tabular}{|>{\raggedright\arraybackslash}m{7.0cm}|m{0.65cm}|}
    \hline
    \textbf{Title} & \textbf{Rank} \\ \hline
   Challenges and Approaches in Multilingual Language Processing & 2 \\ \hline
    Limitations in Dialogue System Quality Assessment and Improvement & 3 \\ \hline
    Reasoning limitations in Large Language Models & 5 \\ \hline
    Advances and Challenges in Summarization Tasks & 6 \\ \hline
    Multi-hop QA Model limitations and Challenges & 7 \\ \hline
    Bias and limitations in AI-driven Hate Speech Detection & 15 \\ \hline
    Gender Bias in Language Models & 21 \\ \hline
  \end{tabular}
  \caption{Some of the Titles of each topic after applying topic modeling (BERTopic + GPT4). A higher rank means more prominent.}
  \Description{Some of the Titles of each topic after applying topic modeling (BERTopic + GPT4). A higher rank means more prominent.}
  \label{tab:title-topic}
\vspace{-20pt}
\end{table}
Table \ref{tab:paper-statistics} shows
the number of papers extracted explicitly using the SciParse tool and the number of papers for which we need a rule-based approach in our dataset. Here, explicit limitation means limitations sections are a separate section, and implicit limitations means limitations sections aren't separate; they are stored as a sub-section or contained inside the other sections.  \\
\textbf{Data Pre-processing:} We removed all punctuation, newline, non-English words, equations, and the string "et al.." We also removed the sentences that contained any links. We removed any limitations section with a word count of less than 15, as we found that the Science Parse tool had extracted unnecessary words. 
Many papers contain irrelevant sentences after the `Limitation' section, such as, "Did you discuss any potential risks of your work?" the science parse tool extracted. So, we checked and removed these types of sentences.
Also, we found that when we extracted `Limitation' sections, it extracted some other section's text. So we checked whether any `limitations' section contained "future work," "ethics," "grants," "appendix," or "section." If we find such a section, we proceed to remove that section from the `limitation' section.

\section{\textsf{Method}}





\begin{figure*}
  \includegraphics[scale=0.87]{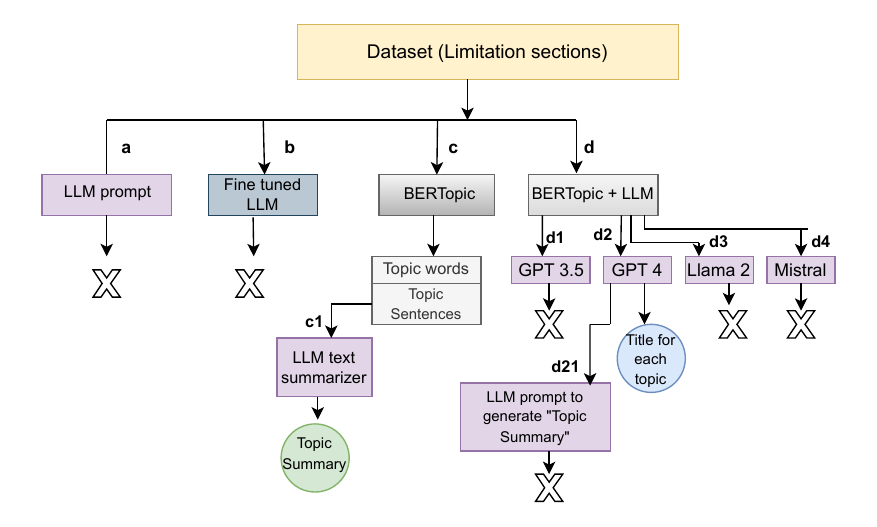}
  \caption{Experiment with various approaches to generate topic titles and Topic Summary. The X mark indicates poor results.} 
  \Description{Experiment with various approaches to generate topic titles and Topic Summary. The X mark indicates poor results.}
  \label{fig:topic-modeling-all-methods}
\end{figure*}

Our proposed model, LimTopic utilizes BERTopic combined with an LLM to analyze the limitations sections of research papers. As shown in Figure 2, BERTopic generates topic words and Topic Sentences for each topic. To enhance clarity, we applied an LLM to generate meaningful titles for each topic and used the LLM again to summarize the Topic Sentences into concise ‘Topic Summaries.

We illustrate our workflow in Figure ~\ref{fig:archi-diagram}. Initially, we gathered ACL papers and extracted the limitations sections $L = {L_1, L_2,....L_n}$ where $n$ is the number of limitations from its corresponding research paper $R = {R_1, R_2,....R_n}$. Then, we used topic modeling on all $n$ limitations sections and generated $j$ topics, denoted by $T_{1}$ to $T_{j}$. Each topic $Topic_{i}$ has a collection of topic words or keywords we can denote as $Topic_{i} = {Tk_1, Tk_2,.....Tk_k}$. Topic modeling uncovers a distribution of words for each identified topic. However, these topics can be hard to understand due to a lack of semantic meaning, leading to poor interpretation of the actual topic. 
Therefore, we used BERTopic along with LLM and generated meaningful title $T_{i}$ and Topic Sentence $R_{i}$ for each topic $Topic_{i}$ (Equation~\ref{eq:topic}). Here $1<=i<=j$. 
Each Topic Sentence $R_{i}$ is very long and hard to interpret, so we use text summarization and generated Topic Summary $TS_{i}$ for each topic $Topic_{i}$ (Equation~\ref{eq:summary}). Hyperameter tuning of our proposed model's LimTopic described in section 5.3 (Figure \ref{fig:topic-modeling-all-methods}d2). 


\begin{equation}
     {T_{1,2,..j} = BERTopic(L_{1,2,..n})} 
      \label{eq:topic}
\end{equation}

\begin{equation}
    {ST_j = LLM summarize(R_j)}
    \label{eq:summary}
\end{equation}



\section{\textsf{Experiments}}

We experimented with four schemes after applying various pre-processing methods to the dataset. \\ 
    \textbf {1. Topic Generation via LLM Prompting:} LLM prompt engineering to generate topics and titles with a Topic summary (Figure \ref{fig:topic-modeling-all-methods}a). \\ 
    \textbf {2. Topic Generation with LLM Fine-tuning:} Utilized fine-tuned Llama 2 to generate topics and titles with a Topic summary (Figure \ref{fig:topic-modeling-all-methods}b). \\
    \textbf{3.Topic Modeling (BERTopic + LLM) with LLM Summarization: Proposed Model (LimTopic)} \\
        a. Integrating BERTopic with LLM as a topic modeling to generate topic, title, and Topic sentences. (Figure \ref{fig:topic-modeling-all-methods}c), (Figure \ref{fig:archi-diagram}, blue box) \\ 
        b. LLM as a text summary on top of topic modeling (BERTopic + LLM) to generate a concise Topic summary (Figure \ref{fig:topic-modeling-all-methods}d), (Figure \ref{fig:archi-diagram}, green box).

\subsection{\textsf{Topic Generation via LLM Prompting}}

Our initial approach used LLM prompt engineering to generate potential topics, titles, and summaries from the dataset containing limitations sections. We leveraged the LangChain library\footnote{https://python.langchain.com/v0.2/docs/introduction/} with OpenAI's API to facilitate prompt engineering. We loaded the entire dataset and used GPT-3.5 with a prompt: “How many topics can you generate by emphasizing limitations and generating topic titles?” GPT-3.5 initially identified around 1000 topics focused on limitations in the dataset. To ensure consistency, we applied the self-consistency approach \cite{wang2022self}, using the same prompt again, which confirmed the generation of 1000 topics. However, upon reviewing these topics, we found them to be repetitive and overly generic, with several topic titles appearing multiple times. We then applied the same approach using GPT-4, which, in contrast, generated only 5 distinct topics, offering more focused and relevant results.
Next, we set 35 as the optimal number of topics, aligning with the 35 topics generated by BERTopic. We prompted GPT-3.5 with: “Can you generate 30-35 topics? Could you create a title and a Topic Summary for each topic within 250 words, with a particular focus on its limitations?” GPT-3.5 generated 30 topics; however, many titles were repetitive, and the summaries were too generic, lacking detailed insights for each topic. Similarly, GPT-4 produced only five topics, failing to capture the nuances of each limitation. To improve performance, we implemented a sequential approach. First, we used a prompt to generate a list of 35 concise topic names (one or two words each), which we referred to as seed words (Table \ref{tab:seed-word}). Using this seed words list, we prompted the model to generate detailed topics, titles, and Topic Summaries based on these seed words, as shown in (Figure \ref{fig:prompt_topic_seed_word}). However, the generated `Topic Summary' is small and lacks sufficient details.


\begin{figure}[ht]
    \centering
    \begin{tikzpicture}
        \node[draw, rectangle, rounded corners, fill=yellow!20, text width=0.45\textwidth, inner sep=5pt, scale=0.9] (box) {
            \begin{minipage}{\textwidth}
                \small
                \textbf{prompt} = \texttt{\textquotesingle\textquotesingle\textquotesingle} \\
                I have this topic list [seed word]. \\
                Can you generate 30-35 topics to try each sentence that belongs to one of these classes? If not, make a new topic. Generate each topic title and Topic Summaryof each topic within 150 words \\
                \texttt{\textquotesingle\textquotesingle\textquotesingle}
            \end{minipage}
        };
    \end{tikzpicture}
    \caption{Prompt for topic generation using seed word}
    \Description{Prompt for topic generation using seed word}
    \label{fig:prompt_topic_seed_word}
    \vspace{-19pt}
\end{figure}







\subsection{\textsf{Topic Generation with LLM Fine-tuning}}
In addition to prompting, our second approach investigated fine-tuning the pre-trained Llama 2 7b model using a 20\% random sample of our dataset and making it a `train dataset.' We made our input as a `Question-Answer' pair where `questions' is the topic title from BERTopic + LLM and `answers' is the text from limitations sections to train the Llama model. We employed a fine-tuned approach with low-rank adaptation (LoRA) \cite{hu2021lora}. LoRA introduces two smaller matrices where the original weight matrix remains frozen during fine-tuning, and these two smaller matrices will add in the original weights. The model will learn new patterns without changing the pre-existing model structure. LoRA helps preserve knowledge by modifying a small fraction of the model's parameters. Then we used QLoRA \cite{dettmers2024qlora} for 4-bit quantization instead of 16 or 32 bits, and it reduced the precision of the numerical values in parameters for speeding up computation.  We set these parameters, `batch size' = 4, `lora attention dimension' = 64, `lora alpha' = 16, and `lora dropout' = 0.1, `maximum token length' = 512, and used the Adam optimizer with a learning rate 2e-4. After 100 iterations, the loss dropped and converged. We fine-tuned the model using the `train dataset' and created a new model and tokenizer. Next, we added test data to the new model and used prompts to generate topics, titles, and Topic Summary for each topic. We approached the prompt using two distinct methods: In our first approach, we used the prompt, ``Based on the above information, extract topics with a short label and Topic Summary." Here, we didn't specify the number of topics. In the second approach, we specified the number of topics to 30-35 and asked prompts to generate titles and Topic Summary for each topic. We chose this number because BERTopic generated this number of topics. However, the fine-tuned Llama model's output was not satisfactory. In both cases, it generated five topics, and the Topic Summary snippets were generic and lacked sufficient detail.



\subsection{\textsf{(LimTopic) Topic Modeling (BERTopic + LLM) with LLM Summarization}}
To alleviate the problem in earlier approaches, such as LLM Prompt ({Figure \ref{fig:topic-modeling-all-methods}a}) and Fine Tuning approach ({ Figure \ref{fig:topic-modeling-all-methods}b}) we implemented BERTopic with LLM ({ Figure \ref{fig:topic-modeling-all-methods}d}), and it shows better performance consisting of better topics, titles, and Topic Summaries. It generates better human-understandable topic titles than ({ Figure \ref{fig:topic-modeling-all-methods}c}) also, which generated topic words or keywords only.

\begin{figure}[ht]
    \centering
    \begin{tikzpicture}
        \node[draw, rectangle, rounded corners, fill=yellow!20, text width=0.45\textwidth, inner sep=5pt, scale=0.9] (box) {
            \begin{minipage}{\textwidth}
                \small
                \textbf{system prompt} = \texttt{\textquotesingle\textquotesingle\textquotesingle} 
                
                You are a helpful, respectful and honest assistant for labeling topics.
                
                \texttt{\textquotesingle\textquotesingle\textquotesingle}
                
                \vspace{2mm}\hrule\vspace{2mm}
                
                \textbf{example prompt} = \texttt{\textquotesingle\textquotesingle\textquotesingle} 
                
                I have a topic that contains the following documents: 
                
                - Traditional diets in most cultures were primarily plant-based with a little meat on top, but with the rise of industrial-style meat production and factory farming, meat has become a staple food. 
                
                - Meat, but especially beef, is the worst food in terms of emissions. 
                
                - Eating meat doesn't make you a bad person, not eating meat doesn't make you a good one. 
                
                The topic is described by the following keyword: `meat, beef, eat, eating, emissions, steak, food, health, processed, chicken'.

                Based on the information about the topic above, please create a short label or title for this topic. Make sure you only return the label and nothing more.

                \texttt{\textquotesingle\textquotesingle\textquotesingle}
                
                \vspace{2mm}\hrule\vspace{2mm}
                
                \textbf{main prompt} = \texttt{\textquotesingle\textquotesingle\textquotesingle} 
                
                I have a topic that contains the following documents: 
                
                [DOCUMENTS] 
                
                The topic is described by the following keyword: [KEYWORD]. 

                Based on the information about the topic above, please create a short label of this topic. Make sure you to only return the label and nothing more. 

                \texttt{\textquotesingle\textquotesingle\textquotesingle}
                
                \vspace{2mm}\hrule\vspace{2mm}
                
                \textbf{prompt} = \texttt{system prompt + example prompt + main prompt}

                \texttt{\textquotesingle\textquotesingle\textquotesingle}
                
                \vspace{2mm}\hrule\vspace{2mm}

                \textbf{Output} \\
                \textbf{Title} = \texttt{Challenges on Dense Retrieval Models in Information Retrieval \\ 
                 \textbf{Topic Sentences}: However, it is unclear how it may be adapted for the single-representation dense retrieval prf model.In addition, in this...  \\ }
            \end{minipage}
        };
    \end{tikzpicture}
    \caption{Prompt for topic modeling (BERTopic + GPT).}
    \Description{Prompt for topic modeling (BERTopic + GPT).}
    \label{fig:prompt_topic_modeling}
   \vspace{-10pt} 
\end{figure}

\subsubsection{\textbf{Experimenting with BERTopic}} We initially experimented with Linear Discriminant Analysis (LDA) for topic modeling, aiming to capture specific limitations for each topic. However, LDA only produces topic words, making it difficult to grasp the actual meaning of each topic. Additionally, the topic summaries generated by LDA often covered unrelated or inconsistent limitations, as LDA grouped diverse sentences that did not always address the same topic coherently. To address these issues, we implemented BERTopic, which, unlike LDA, gathers related sentences for each topic as Topic Sentences, providing a more cohesive context. While BERTopic generates topic words, determining a clear title for each topic remains challenging. To improve interpretability, we integrated an LLM with BERTopic to generate descriptive titles (Figure \ref{fig:archi-diagram}, blue area). Furthermore, we applied LLM-based text summarization to refine Topic Sentences into focused summaries that clearly convey each topic’s specific limitations (Figure \ref{fig:archi-diagram}, green area).
We began experimenting with BERTopic using a pre-processed dataset (Figure \ref{fig:BERTopic}). BERTopic has various elements where the tokenizer converts text into smaller tokens, UMAP \cite{mcinnes2018umap} reduces the dimension of text embedding for clustering, Count Vectorizer creates a matrix of token counts which calculates class-based TF-IDF score, and Sentence Transformers create dense embedding vectors with capturing semantic meanings of the sentence. Figure \ref{fig:BERTopic} dotted line denotes those elements are optional in BERTopic. We experimented with those optional elements and found that integrating `seed words' (Table \ref{tab:seed-word}) improves performance. We explored combinations of these approaches, such as: \\
a) with/without seed word, b) with/without UMAP and c) with/without HDBSCAN. Here, the seed word contains the prominent word for each topic in the dataset. We use these seed words for guided topic modeling. To generate seed words, we used a GPT-4 prompt to produce 30-35 topic names, each consisting of one or two words (Table \ref{tab:seed-word}). We then used this list of words to guide topic modeling in BERTopic. We use UMAP for dimensionality reduction, which involves capturing lower dimensions from local and global high-dimensional space. 
We also experimented with HDBSCAN in BERTopic. However, when employing UMAP, it copies the example of the prompt and produces it in the topic summary for some topics.
After reducing the dimensionality of input embeddings, we use HDBSCAN \cite{mcinnes2017hdbscan}, a clustering process, to extract topics. However, HDBSCAN decreased the performance. One reason is that BERTopic generated high-dimensional data, whereas HDBSCAN used Euclidean distance, which doesn't capture the semantic distance between words. We experimented with the combination of UMAP, HDBSCAN, and seed word and measured the performance of the silhouette and coherence scores as a quantitative study. Only including seed words with BERTopic improved performance, whereas UMAP and HDBSCAN led to decreased coherence and silhouette scores. BERTopic generated topic words or keywords for each topic. However, LLM can generate more coherent and descriptive titles for each topic, making more straightforward interpretability, and deeper contextual understanding can capture the breadth and nuances of a topic. So, we integrate LLM with BERTopic to create titles for each topic.  



\subsubsection{\textbf{Prompt Engineering on LLM for topic modeling}} 

 
After applying BERTopic, we used a large language model (LLM) to generate topic titles. BERTopic allows customization through various parameters, such as text embedding, dimensionality reduction (using UMAP), clustering (using HDBSCAN), and representation. In this case, we used the LLM as the representation model, illustrated by the label ‘Rep. Model’ in Figure \ref{fig:BERTopic}. By using LLM-generated outputs for topic modeling, we improved the quality of topic titles, moving beyond simple keywords to meaningful, refined titles. Additionally, the LLM refined the textual output provided by BERTopic. For this process, we designed two types of prompts for the LLM. In the first approach, we used prompts to generate concise topic titles (shown in Figure \ref{fig:prompt_topic_modeling}). In the second approach, we used prompts that created both topic titles and a brief Topic Summary for each topic (Figure \ref{fig:prompt_topic_generation_related_text}). While these summaries gave an overview of each topic, they were intentionally kept general and did not generate detailed content. On the other hand, BERTopic also generated more comprehensive 'Topic Sentences' by aggregating content from relevant sections throughout the text (Figure \ref{fig:archi-diagram}, blue area). These Topic Sentences were lengthy and detailed, covering nuanced aspects of each topic. To condense this information, we applied a text summarization technique with the LLM (section 5.3.3) to create clearer, focused Topic Summaries for each topic (Figure \ref{fig:archi-diagram}, green area). We also experimented with few-shot and zero-shot prompting techniques \cite{kojima2022large}. Testing with one to five examples for few-shot learning, we found that providing one example worked best; using more examples sometimes led to overfitting, where the LLM would incorporate content from the example responses directly into the summaries, reducing the quality of the output.


\begin{figure}[ht]
    \centering
    \begin{tikzpicture}
        \node[draw, rectangle, rounded corners, fill=yellow!20, text width=0.4\textwidth, inner sep=5pt, scale=1.0] (box) {
            \begin{minipage}{\textwidth}
                \small
                \textbf{prompt} = \texttt{\textquotesingle\textquotesingle\textquotesingle} \\
                I have a topic that contains the following documents: [DOCUMENTS]. \\
                The topic is described by the following keyword: [KEYWORD]. \\
                Based on the information above, extract a short topic label and `Topic Summary' in the following format: topic: <topic label> Topic Summary: <texts>" \\
                \texttt{\textquotesingle\textquotesingle\textquotesingle}
                
                \vspace{2mm}\hrule\vspace{2mm}
                \textbf{Output} \\
                \textbf{Title} = \texttt{Challenges and limitations in Clinical Data Usage and NLP Model Evaluation in Healthcare \\ 
                 \textbf{Topic Summary}: In this text elucidates significant constraints encountered in the use of clinical data for NLP model evaluation, particularly in healthcare settings. The study’s...  \\ }
                
            \end{minipage}
        };
    \end{tikzpicture}
    \caption{Prompt for topic generations with Topic Summary.}    
    \Description{Prompt for topic generations with Topic Summary.}
    \label{fig:prompt_topic_generation_related_text}
\end{figure}




\subsubsection{\textbf{Experimenting on LLM for text summarization}}
In the previous approach, we didn't get a good `Topic Summary' for each topic using the prompt in BERTopic + LLM (section 5.3.2, Figure: \ref{fig:prompt_topic_generation_related_text}, and BERTopic only (section 5.3.1). BERTopic generated topic sentences for each topic, which collects various texts related to the topic across the dataset. Each topic's Topic Sentences are long, averaging 800 words, making them lengthy and potentially encompassing multiple limitation types. 
So, we started with Topic Sentences for each topic. To address this, we applied LLM-based summarization for each topic's Topic Sentences by employing the prompt in Figure \ref{fig:prompt_text_summarization} and generated a Topic summary for each topic.

\begin{figure}[ht]
    \centering
    \begin{tikzpicture}
        \node[draw, rectangle, rounded corners, fill=yellow!20, text width=0.4\textwidth, inner sep=5pt, scale=1.0] (box) {
            \begin{minipage}{\textwidth}
                \small
                \textbf{prompt} = \texttt{\textquotesingle\textquotesingle\textquotesingle} \\
                "Can you summarize the following texts within 130-140 words, putting more emphasis on the limitations of the `title'?"
                \texttt{\textquotesingle\textquotesingle\textquotesingle}
            \end{minipage}
        };
    \end{tikzpicture}
    \caption{Prompt for text summarization.}
    \Description{Prompt for text summarization.}
    \label{fig:prompt_text_summarization}
\end{figure}

This process yielded concise summaries offering a generalized overview of each topic's specific limitations (See Tables   \ref{tab:title-topic}, \ref{tab:summary-various-llm-judge-gpt4}). 
We also experimented with various LLMs for summarization, including GPT 4, GPT 3.5, Llama 7B, Llama 13B, and Mistral, and used GPT 4 and Claude 3 as a Judge.

\begin{table}[h!t]
  \centering
  \begin{tabular}{|>{\raggedright\arraybackslash}m{4.2cm}|m{3.0cm}|}
    \hline
    \textbf{Category} & \textbf{Optimal parameters} \\ \hline
    number of neighbors (UMAP) & 13 \\ \hline
    number of components (UMAP) & 7 \\ \hline
    min topic size & 10  \\ \hline
    zero shot min similarity & 0.75 \\ \hline
  \end{tabular}
  \caption{Optimal parameters for BERTopic.}
  \Description{Optimal parameters for BERTopic.}
  \label{tab:Optimal-parameters}
  \vspace{-20pt}
\end{table}

\begin{table*}[htbp]
  \centering
  \small 
  \begin{tabular}{l|c|c|c|c}
    \hline
    \multirow{2}{*}{\textbf{Model Name}} &
    \multicolumn{2}{c|}{\textbf{Topic and Title}} &
    \multicolumn{2}{c}{\textbf{Topic, Title with Topic Summary}} \\ \cline{2-5}
    & \textbf{Silhouette Score} & \textbf{Coherence Score} & \textbf{Silhouette Score} & \textbf{Coherence Score} \\ \hline
    LDA & - & - & - & 0.375 \\ \hline
    BERTopic & - & - & 0.577 & 0.601 \\ \hline
    BERTopic + UMAP & - & - & 0.586 & 0.581 \\ \hline
    BERTopic + GPT 3.5 (zero-shot) & 0.551 & 0.468 & 0.544 & {0.520} \\ \hline
    BERTopic + GPT 3.5 (few-shot) & 0.407 & - & 0.546 & 0.427 \\ \hline
    BERTopic + GPT 4 (zero-shot) & 0.561 & 0.487 & 0.529 & 0.596 \\ \hline
    \textbf{BERTopic + GPT 4 (few-shot)} & 0.530 & 0.509 & \textbf{0.588} & \textbf{0.617} \\ \hline
    BERTopic + Llama 7b (zero-shot) & - & - & 0.576 & 0.497 \\ \hline
    BERTopic + Llama 7b (few-shot) & - & - & 0.558 & 0.494 \\ \hline
    BERTopic + Llama 13b (zero-shot) & - & - & 0.578 & 0.517 \\ \hline
    BERTopic + Llama 13b (few-shot) & - & - & 0.558 & 0.519 \\ \hline
    BERTopic + Mistral 7b (zero-shot) & - & - & 0.524 & 0.551 \\ \hline
  \end{tabular}
  \caption{Performance comparison of different models for Topic Modeling.}
  \Description{Performance comparison of different models for Topic Modeling.}
  \label{tab:performance-different-models}
  \vspace{-10pt}
\end{table*}

\begin{table*}[htbp]
  \centering
  \small
  \begin{tabular}{l|c|c|c|c}
    \hline
    \textbf{Sentence Transformer} & \textbf{Min Topic Size} & \textbf{Clusters} & \textbf{Silhouette Score} & \textbf{Coherence Score} \\ \hline
    Bge-base-en-v1.5 & 10 & 23 & 0.542 & 0.632 \\ \hline
    allenai-specter & 10 & 30 & 0.493 & 0.608 \\ \hline
    paraphrase-MiniLM-L6-v2 & 10 & 28 & 0.502 & 0.544 \\ \hline
    all-mpnet-base-v2 & 10 & 39 & 0.590 & 0.591 \\ \hline
    all-MiniLM-L12-v2 & 10 & 33 & \textbf{0.601} & 0.581 \\ \hline
    all-MiniLM-L12-v2 (without umap) & 10 & 32 & 0.578 & 0.601 \\ \hline
    \textbf{all-MiniLM-L6-v2} & \textbf{10} & \textbf{31} & 0.528 & \textbf{0.715} \\ \hline
    paraphrase-multilingual-MiniLM-L12-v2 & 10 & 28 & 0.460 & 0.521 \\ \hline
    paraphrase-MiniLM-L6-v2 & 10 & 28 & 0.502 & 0.544 \\ \hline
    bert-base-nli-mean-tokens & 10 & 2 & 0.454 & 0.503 \\ \hline
    all-distilroberta-v1 & 10 & 34 & 0.591 & 0.594 \\ \hline
    msmarco-distilbert-dot-v5 & 10 & 30 & 0.460 & 0.548 \\ \hline
    multi-qa-MiniLM-L6-cos-v1 & 10 & 25 & 0.498 & 0.627 \\ \hline
  \end{tabular}
  \caption{Performance of various sentence transformers for topic modeling in BERTopic.}
  \Description{Performance of various sentence transformers for topic modeling in BERTopic.}
  \label{tab:Sentence-Transformer}
  \vspace{-10pt}
\end{table*}

\section{\textsf{Results and Analysis}}
We used coherence and silhouette scores to measure the quality of topic modeling. The silhouette score, ranging from -1 to 1, evaluates how well each cluster is formed by comparing intra-cluster similarity with inter-cluster dissimilarity. The coherence score assesses semantic similarity among frequently occurring words within a topic, indicating how well the words and phrases align within a topic’s context. For measuring the quality of text summarization, we used ROUGE-1, ROUGE-2, ROUGE-L, BLEU, and BERTScore as a quantitative method \ref{tab:text-summarizer-bertscore}, and some other metrics such as Grammatically, Readability, so on as a LLM as a judge method (Table \ref{tab:llm-judge}). 



\begin{table*}
  \centering
  \begin{tabular}{l}
      \hline
      You are a very helpful, respectful assistant. Please take time to fully read and understand the \\ context.
      Please select which "Prompt" is better in terms of Grammaticality, Cohesiveness, Understandability, \\ Likability, Cohesiveness, Coherence, Likability, Relevance, Fluency, and describing a very good way of a particular \\ limitations. Note: Please take the time to read and understand the story fragment fully. Rate each prompt out \\
      of 5 ratings, where 5 is best and 1 is worst.
      
  \end{tabular}
  \caption{Prompt for GPT 4 to evaluate summary.}
  \Description{Prompt for GPT 4 to evaluate summary.}
  \label{tab:prompt-GPT4}
  \vspace{-10pt}
\end{table*}


\subsection{\textsf{Topic Modeling:}}  

\begin{table*}[htbp]
  \centering
  \small 
  \begin{tabular}{l|c|c|c|c}
    \hline
    Language & Dataset & Corpus Size & Computational & Machine Translation \\ \hline
    Tokenization & Costs & Interpretability & Morphology & Semantic \\ \hline
    Memory & Skewed Distributions & Biased Distributions & Small Size & Generalizability \\ \hline
    Bias & Hyperparameter & Hardware & Real world & Robustness \\ \hline
    Noisy & Time & Annotations & Evaluation & Diversity \\ \hline
    Segmentation & Metrics & - & - & - \\ \hline
  \end{tabular}
  \caption{Seed words: these words generated by LLM from the dataset are used as a chain approach for topic generation.}
  \Description{Seed words: these words generated by LLM from the dataset are used as a chain approach for topic generation.}
  \label{tab:seed-word}
  \vspace{-10pt}
\end{table*}


\begin{table}[ht]
  \centering
  \begin{tabular}{|>{\centering\arraybackslash}m{1.3cm}|>{\centering\arraybackslash}m{1.3cm}|>{\centering\arraybackslash}m{1.3cm}|>{\centering\arraybackslash}m{1.3cm}|>{\centering\arraybackslash}m{1.3cm}|}
    \hline
   \textbf{Model} & \textbf{GPT 4 Turbo} & \textbf{GPT 4} & \textbf{GPT 3.5 Turbo} & \textbf{GPT 3.5} \\ \hline
    \textbf{Silhouette Score} & \textbf{0.588} & 0.518 & 0.546 & 0.540 \\ \hline
  \end{tabular}
  \caption{Topic modeling performance comparison in GPT variants with BERTopic.}
  \Description{Topic modeling performance comparison in GPT variants with BERTopic.}
  \label{tab:gpt-judge}
  \vspace{-19pt}
\end{table}


BERTopic generates topic words for which we can easily measure the coherence score.  However, measuring the coherence score using a title instead of topic words is impossible. To alleviate this problem, we applied keyBERT to each topic's Topic Sentences and extracted keywords using BERT-embedding and cosine similarity, which were the most similar to the documents. Next, we calculate the average coherence score for each topic (Table \ref{tab:performance-different-models}). 
Table \ref{tab:performance-different-models} shows several topic-modeling models we used; some are traditional topic models like LDA and BERTopic. We start the experiment with LDA, a probabilistic model with a 0.375 coherence score, whereas BERTopic gives a big jump to a 0.601 coherence score, and if we use UMAP, the coherence score drops. UMAP is one approach used for dimensionality reduction to speed up the training process, but its drawback is that it requires assistance handling large amounts of data. Then, we added LLM with BERTopic to generate titles for each topic. We applied two prompts: one generates a `Topic and title,' and the other generates a `Topic, title with the Topic Summary.' We found that the latter approach showed better coherence and silhouette scores when we applied BERTopic with GPT 4. We also experimented with a few shots and a zero-shot approach, and a few shots achieved a 0.021 more coherence score and 0.059 more silhouette score. Also, BERTopic + GPT 4 (few shots) achieves 0.016, 0.19, 0.123, 0.098 more performance than BERTopic, BERTopic + GPT 3.5 (few shots), Llama 7b (few shots), Llama 13b (few shots)  in terms of coherence score respectively.

Our proposed BERTopic + GPT-4 model produced 35 topics with descriptive titles with the rank. Here, a higher rank means more prominent this type of `Limitation' in the dataset. (See Table \ref{tab:title-topic}). 

In GPT 4, we experimented with one, two, three, four, and five examples in a few shots and found that one shot is better, whereas in other cases, the generated text mimics the example from the prompt. The performance of LLM in producing subject titles from the input text is highly contingent on the size of the context window. We used GPT 3.5 Turbo, which has a context window size of 16,385 tokens, whereas GPT 4 Turbo has a larger context window of 128,000 tokens. Also, BERTopic + GPT4 has the capability to achieve a stronger understanding with more parameters trained on vast datasets, capturing the abstract themes of keywords provided by BERTopic, which ends up with GPT 4 outperforms other models (Table \ref{tab:performance-different-models}).

In other cases, the output was not sound when we integrated BERTopic with GPT 3.5, Llama 7b, Llama 13b, or Zephyr (Mistral 7b). Here, Llama 7b, Llama 13 b, and Mistal generated topic word (keyword) for each topic. We used a prompt to create titles here, but the topic titles must be more concise. GPT 3.5 produced very little Topic Summary; Llama 7b and 13b produced some Topic Summary on certain topics; and Zephyr (Mistral 7b) also produced some irrelevant topic titles and Topic Summarys. One possible reason for the poor performance of Zephyr (Mistral 7b) is that it's got 512 tokens as a context length when we integrate with BERTopic, where the input length is much higher (Table \ref{tab:performance-different-models}, Fig: \ref{fig:Dataset_size_vs_Performance}). Llama 7b, 13b produces some irrelevant topics and titles because of the smaller parameter size and context window compared to GPT. 

We also experimented with GPT 4 and GPT 3.5 variants with BERTopic for topic modeling. Both models have famous variants, such as GPT 4/3.5 and GPT 4/3.5 turbo. We found that the GPT turbo version performs better than the GPT version for topic modeling in Table \ref{tab:gpt-judge}.  Here `gpt-4-turbo-preview' performs best (Tables \ref{tab:performance-different-models}, \ref{tab:gpt-judge}). We found that `all-MiniLM-L6-v2' achieves the best coherence score, and `all-MiniLM-L12-v2' achieves the best silhouette score (Table \ref{tab:Sentence-Transformer}). One reason is that those embeddings are fine-tuned for general purposes, capable of preserving key semantic information for embedding, and trained on diverse datasets. Another important parameter is `minimum topic size,' which specifies a topic's minimum size. We experimented with minimum topic sizes of 7, 10, and 12 and found that 10 is optimal. The minimum topic size of less than 10 will generate many overlapping topics, and more than 10 will generate fewer topics by combining two or more topics. In addition, we experimented with another parameter, `zero-shot min similarity,' experimenting with 0.7, 0.75, 0.8, and 0.85 and found 0.75 provides the best coherence score. Additionally, we used `CountVectorizer' and removed the stopwords in the pre-processing stage. Furthermore, adding more data enhances the quality of topic word in the BERTopic + LLM (GPT 4, GPT 3.5, Llama 2), ensuring that each topic's word aligns well with their respective cluster. We also experimented using BERTopic with UMAP. After varying the number of neighbors from 1 to 30, we found that 13 is the optimal number of neighbors, and 7 is the optimal number of components (Table \ref{tab:Optimal-parameters}). We integrated optimized various parameters, such as Sentence Transformer, Count Vectorizer, UMAP, Seed words, Tokenzier, GPT 4 Prompt \footnote{https://maartengr.github.io/BERTopic} (Figure \ref{fig:prompt_topic_modeling}), Document Length, Embedding Model, and Minimum Topic Size with GPT 4 with BERTopic. We also analyzed each category of ACL 2023 papers to identify prominent limitations within each category. These limitations are in Table \ref{tab:diff-categories-lim}.



\subsection{\textsf{Text summarization:}} After applying topic modeling with BERTopic + LLM, BERTopic generated Topic Sentences. We applied LLM as a text summarizer to summarize each topic's Topic Sentences concisely as we depict it as a Topic Summary (see Table \ref{tab:summary-various-llm-judge-gpt4}). We applied GPT 4, Claude 3.5 Sonnet, Llama 3, and GPT 3.5 to summarize each topic Topic Summary. At first, we used ROUGE \cite{lin2004rouge} score to measure the performance of each LLM as a summarizer. We used Topic Sentences generated by BERTopic from the dataset, containing texts of each topic. We took these Topic Sentences as a reference text and considered the summary from GPT 4, Claude 3.5 Sonnet, Llama 3, and GPT 3.5 as candidates. We check the ROUGE-1, ROUGE-2, and ROUGE-L scores, which show the overlaps of unigrams, bigrams, and the longest common subsequence between the candidate summary and the reference text. BLEU measures the number of matching words and phrases between LLM-generated text and reference text with an overlap of different n-gram lengths. 


\begin{figure}
 \includegraphics[scale=0.35]{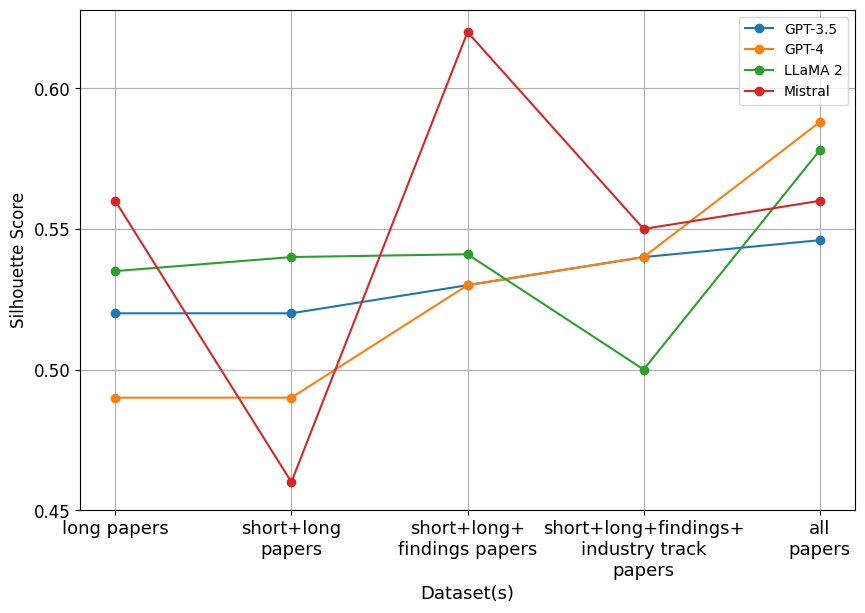}

  \caption{Dataset size vs Performance}
  \Description{Dataset size vs Performance}
  \label{fig:Dataset_size_vs_Performance}
    \vspace{-15pt}

\end{figure}



In some cases, the GPT 4 and Claude 3.5 Sonnet summary ratings are similar, providing detailed, cohesive, and fluent discussion. However, GPT 4 maintains a high level of detail without sacrificing technical depth. We found that the summary from GPT 3.5 achieves the highest ROUGE-1 score, ROUGE-2, ROUGE-L score, and BLEU score. So, a higher value means more overlapping words between actual and LLM-generated limitations. A greater degree of overlap means copying the word from the actual limitations as a summarizer, which is not good. The problem with these metrics is that they rely on exact matches between the LLM summary and reference text. So, we applied another metric named BERTScore, which relies on semantic similarity between generated and reference text tokens. BERTScore has precision, recall, and F1 scores, where recall measures how much the content in the candidate sentence captures the essence of the reference using cosine similarity. So, we considered `recall' in BERTScore as we used Topic Sentences as a `reference text,' which consists of the actual limitations of each topic. Table \ref{tab:text-summarizer-bertscore} shows GPT 4 outperforms BERTScore (recall) compared to Claude 3.5 Sonnet, Llama 3, and GPT 3.5. 

To improve the quality of summaries, we opted to use human-generated summaries as ‘reference text’ or ground truth rather than relying solely on text directly from the paper. To further enhance this evaluation, we implemented an alternative method called LLM judgment, where GPT-4 and Claude 3.5 Sonnet served as evaluators \cite{zheng2024judging}. First, we generated summaries using GPT-4, LLaMA 3, Claude 3.5 Sonnet, and GPT-3.5, and then asked GPT-4 to rate each summary on a scale from 1 to 5. The rating criteria included grammar, readability, cohesiveness, understandability, likability, coherence, relevance, fluency, and overall description quality (Table \ref{tab:llm-judge}, \ref{tab:summary-various-llm-judge-gpt4}). We repeated this process using Claude 3.5 Sonnet as the evaluator, asking it to rate summaries from all the LLMs using the same metrics (Table \ref{tab:prompt-GPT4}). From the set of 35 summaries, we randomly selected 15 and calculated the average scores from both GPT-4 and Claude 3.5 Sonnet. Table (\ref{tab:summary-various-llm-judge-gpt4}) provides a comparison of summaries generated by various LLMs along with their average ratings. Overall, GPT-4, as a summarizer, consistently received higher ratings than the other LLMs, indicating its superior performance (Table \ref{tab:llm-judge}).

\begin{table}[ht]
  \centering
    \small
  \begin{tabular}{|>{\raggedright\arraybackslash}m{2cm}|m{0.9cm}|m{0.95cm}|m{1.2cm}|m{0.9cm}|}
     \hline
    \textbf{Metrics} & \textbf{GPT 4} & \textbf{Llama 3} & \textbf{\makecell{Claude 3.5 }} & \textbf{GPT 3.5} \\ \hline
    Grammatically & \textbf{4.75} & 4.15 & 4.57 & 3.6 \\ \hline
    Readability & \textbf{4.11} & 3.6 & 4 & 2.95 \\ \hline
    Cohesiveness & \textbf{4.75} & 3.55 & 4.3 & 2.8 \\ \hline
    Understandability & \textbf{4.11} & 3.65 & 4.05 & 2.8 \\ \hline
    Likability & \textbf{4.22} & 3.4 & 4 & 2.85 \\ \hline
    Coherence & \textbf{4.75} & 3.55 & 4.3 & 2.75 \\ \hline
    Relevance & \textbf{4.81} & 3.95 & 4.55 & 3.15 \\ \hline
    Fluency & \textbf{4.36} & 3.65 & 4.2 & 2.8 \\ \hline
    Description Quality & \textbf{4.8} & 3.55 & 4.45 & 2.7 \\ \hline
  \end{tabular}
  \caption{Performance of summarization in various LLM judged by GPT 4 and Claude 3.5 Sonnet.}
  \Description{Performance of summarization in various LLM judged by GPT 4 and Claude 3.5 Sonnet.}
  \label{tab:llm-judge} 
    \vspace{-23pt}
\end{table}


\begin{table}[htbp]
  \centering
  \small
   \begin{tabular}{|m{1.1cm}|m{1cm}|m{1cm}|m{1cm}|m{0.75cm}|m{1.25cm}|} 
    \hline
    \textbf{Model} & \textbf{ROUGE-1} & \textbf{ROUGE-2} & \textbf{ROUGE-L} & \textbf{BLEU} & \textbf{BERTScore} \\ \hline
    GPT 4 & 34.13 & 9.34 & 16.95 & 5.18 & \textbf{83.61} \\ \hline
    Claude 3.5 Sonnet & 36.18 & 10.5 & 15.06 & 4.55 & 83.31 \\ \hline
    Llama 3 & 45.34 & 16.96 & 16.38 & 3.86 & 82.47 \\ \hline
    GPT 3.5 & \textbf{62.26} & \textbf{28.98} & \textbf{25.74} & \textbf{6.42} & 82.46 \\ \hline
  \end{tabular}
  \caption{Performance of various LLM as a text summarizer. Input is Topic Sentences from BERTopic.}
  \Description{Performance of various LLM as a text summarizer. Input is Topic Sentences from BERTopic.}
  \label{tab:text-summarizer-bertscore}
    \vspace{-19pt}
\end{table}


Fig \ref{fig:Dataset_size_vs_Performance} depicts that increasing the dataset size also increases the silhouette score. Here, we started experimenting with one category dataset, which contains 'long papers', then we gradually added other category datasets such as `short papers,' `findings ACL papers, `industry track papers,' and lastly, we included all 28 categories, and we found silhouette score is increasing in GPT 3.5, GPT 4, and Llama 2. A better silhouette score means the topic quality (cluster) is good. With the comparison of four LLMs, GPT 3.5 achieves a 0.52 score, gradually increasing to 0.54 after adding each category dataset. Llama 2 is slightly inconsistent, where its performance drops at one point and increases later. The exception occurred by Mistral because the token limit is 512 tokens, and our dataset exceeds the limits. GPT 4, from 0.49 to 0.58, performs very well and is stable compared to GPT 3.5, Llama 2, and Mistral, where the performance increases after adding each dataset category. GPT 4 makes outstanding topics if we provide a sufficient number of datasets.

\begin{table}[htbp]
  \centering
  \begin{tabular}{|p{6.4cm}|p{1.2cm}|}
    \hline
    \textbf{Title:} Challenges and limitations in Clinical Data Usage and NLP Model Evaluation in Healthcare & \textbf{Results}  \\ \hline
    
    \textbf{GPT 4 summary}: In this text elucidates significant constraints encountered in the use of clinical data for NLP model evaluation, particularly in healthcare settings. The study's reliance on a single clinical cohort limits its generalizability, raising concerns about the representativeness of the data due to possible biases in gender, age, race distributions. While promising for cancer immunotherapy prognostic prediction, the study's findings may not necessarily apply to other clinical trials without further validation...   &
    Gramm: 5, 
    Readability: 4,
    Cohesiveness: 5, 
    Understandability: 4, 
... 
 \\
    \hline
  \end{tabular}
  \caption{Summary from GPT 4 and judge by GPT 4 and Claude 3.5 Sonnet.}
  \label{tab:summary-various-llm-judge-gpt4}
  \vspace{-20pt}
\end{table}

\begin{table}[h!t]
  \centering
  \begin{tabular}{|>{\raggedright\arraybackslash}m{5.7cm}|m{1.6cm}|}
    \hline
    \textbf{Limitation} & \textbf{Category} \\ \hline
    Enhancing medical summarization and model reliability.
 & Medical NLP \\ \hline
    Limitations and challenges in evaluating conversations with LLMs. & NLP conv AI \\ \hline
    Speech-to-text translation challenges. & IWSLT  \\ \hline
    Dialogue system development and evaluation.
 & Finding ACL \\ \hline
  \end{tabular}
  \caption{Limitation in some categories of ACL 2023 papers.}
  \Description{Limitation in some categories of ACL 2023 papers.}
  \label{tab:diff-categories-lim}
  \vspace{-20pt}
\end{table}


\section {\textsf{Conclusion}}
 
In this paper, LimTopic used BERTopic and GPT 4 for topic modeling to analyze sections with limitations from scientific articles, creating titles and topics with Topic Sentences. Later, we applied text summarization to each topic's Topic Sentences to generate a concise Topic summary for each topic. 
We tried three approaches (Figure \ref{fig:topic-modeling-all-methods}); firstly, we used our dataset and applied the LLM prompt to generate titles and Topic Summarys. Secondly, we applied LLM fine-tuning to create titles and Topic Summarys; we tried to prompt in BERTopic + LLM to generate Topic Summary. Here, BERTopic + GPT 4 generates a perfect title, and all the above methods generate concise and generic Topic Summaries. We applied text summarization of each topic's Topic Sentences to create a Topic summary. 
Here, BERTopic created Topic Sentences that consist of various sentences pertinent to each topic from all over the dataset.  So we took the Topic Sentences, applied various LLM for summarization tasks, and created a better Topic summary for each topic. We also measured the performance of each summary from LLM using Chatgpt 4 and Claude 3.5 Sonnet. 
Our approach not only depicts the limitations but also suggests a structure to address the current limitations. It promotes a thorough and critical approach to future research and a nuanced interpretation of existing works. Furthermore, beyond LLMs' typical language processing ability, we evaluate several LLMs to do complex analytical tasks such as topic modeling.









\section*{\textsf{Limitations and Future Work}}
In this work, we focused only on LLM-based topic modeling to generate topics and used a text summarization approach with LLMs to facilitate understanding of each topic's details. Due to API costs, large-scale analysis with GPT could have been more feasible. Additionally, GPU limitations restricted experimentation to Llama 7b and Llama 13b models. Moreover, our analysis was limited to the ACL 2023 dataset, which might affect the generalizability of findings and some potential biases, such as ACL 2023 papers reflecting the current methodologies and trends (e.g., LLMs), which overrepresents cutting-edge research that might not be relevant to earlier research. This dataset also focuses on particular areas that researchers are interested in but might not represent the overall scenario. Our model shows a methodological bias toward prominent topics and titles, which leads to the underrepresentation of alternative methodologies. We rely on the LLMs to generate topic modeling and text summarization, which might create bias from the dataset on which LLMs are trained. Also, the GPT model trained on a specific date and didn't include information after their last training data update. LLM might have certain biases, such as language bias and demographic bias. 
Future work will also incorporate Retrieval Augmented Generation (RAG) with LLM, fine-tuning with Llama 3, ensuring different viewpoints, with more diverse dataset, and utilizing an iterative approach with a human feedback loop to alleviate these problems. We will extend to diverse datasets from various venues over a broader timeframe with different LLM fine-tuning approaches. We will take the human-generated summary as a ground truth to evaluate text summarization. We will extend our work using qualitative analysis for the performance of LLMs for text summarization.

\bibliographystyle{ACM-Reference-Format}
\bibliography{sample-base}


\label{sec:appendix}

\end{document}